%% file: emnlp2022.tex
\pdfoutput=1

\documentclass[11pt]{article}

\usepackage{EMNLP2022}

\usepackage{times}
\usepackage{latexsym}

\usepackage[T1]{fontenc}

\usepackage[utf8]{inputenc}

\usepackage{microtype}

\usepackage{inconsolata}

\usepackage{booktabs}
\usepackage{multirow}
\usepackage{tikz}
\usepackage{amssymb}
\usepackage{amsmath}

\title{{CUNI} Non-Autoregressive System for the {WMT}~22 Efficient Translation Shared Task}

\author{Jindřich Helcl \\
  Charles University, Faculty of Mathematics and Physics \\
  Institute of Formal and Applied Linguistics \\
  Malostranské náměstí 25, 11800 Prague, Czech Republic \\
  \texttt{helcl@ufal.mff.cuni.cz}}

\begin{document}
\maketitle
\begin{abstract}
We present a non-autoregressive system submission to the {WMT}~22 Efficient Translation Shared Task. Our system was used by \citet{helcl-etal-2022-non}
in an attempt to provide fair comparison between non-autoregressive and autoregressive models. This submission is an effort to 
establish solid baselines along with sound evaluation methodology, 
particularly in terms of measuring the decoding speed. The model itself
is a 12-layer Transformer model trained with connectionist temporal classification on knowledge-distilled dataset by a strong autoregressive teacher model.
\end{abstract}

\section{Introduction}

In the past few years, non-autoregressive (NAR) models for neural machine translation (NMT) attracted interest from the research community \citep{gu2017nonautoregressive,lee-etal-2018-deterministic}. Given the conditional independence between the output states, the decoding process
can be parallelized across time steps. In theory, this leads to higher
decoding speeds.

Since efficient decoding is claimed to be the main motivation of
non-autoregressive models, the Efficient Translation Shared Task seems
to be the appropriate venue to provide fair comparison between these models and their autoregressive counterparts. However, all submissions to this
task were autoregressive so far \citep{birch-etal-2018-findings,hayashi-etal-2019-findings,heafield-etal-2020-findings,heafield-etal-2021-findings}.

Recently, \citet{helcl-etal-2022-non} pointed out common flaws in the evaluation methodology of NAR models. We found that optimized autoregressive models still achieve superior performance over NAR models. The only scenario 
where NAR models showed some potential is GPU decoding with batch size of 1 (latency). Nevertheless, optimized autoregressive models were still both faster and better in terms of translation quality. The main purpose of this submission is to provide a reasonable baseline to future non-autoregressive submissions.

\section{Model}
In our experiments, we use the non-autoregressive model proposed by \citet{libovicky-helcl-2018-end} based on Connectionist Temporal Classification (CTC; \citealp{graves2006connectionist}). We submit models that have been trained as a part of \citet{dizertace}.

\paragraph{Architecture.} The architecture is a 6-layer Transformer encoder, followed by a state-splitting layer and another stack of 6 Transformer layers.
The state-splitting layer takes the encoder states, project them into $k$-times wider states using an affine transformation, and then split the states
into $k$-times longer sequence while retaining the original model dimension.
In the submitted model, we set $k=3$. The latter 6 layers cross-attend to the states immediately after state-splitting. We use Transformer model dimension of 1,024, 16 attention heads and a dimension of 4,096 in the feed-forward sublayer.  

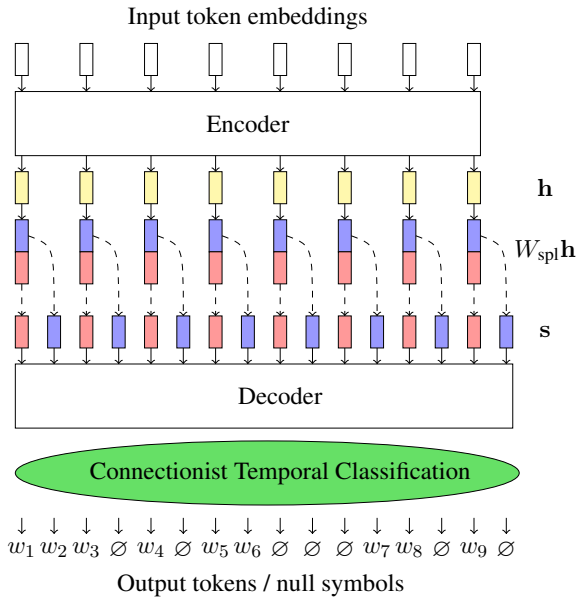
\begin{figure}
  \centering
  \scalebox{0.85}{\input{img/non-autoregressive}}
  \caption{The CTC-based model architecture. We show
    the original image from \citet{libovicky-helcl-2018-end}.}
  \label{fig:ctc-schema}
\end{figure}

The defining property of non-autoregressive models is that the decoding process treats output states as conditionally independent. In this architecture, we set the output sequence length to $k\times T_x$ where $T_x$ is the length of the source sentence. To allow for shorter output sequences, the any output state can produce an empty token. The training loss is then
computed using a dynamic programming algorithm as a sum of losses of all possible empty token alignments which lead to the same output sentence. The schema of the architecture is shown in Figure ~\ref{fig:ctc-schema}.

\paragraph{Training.} We train our model on the knowledge-distilled data generated by the provided teacher \citep{chen-etal-2021-university}. We use learning rate of 0.0001 in a inverse square-root decay scheme with 8,000 warm-up and decay steps.

\paragraph{Implementation.} We implement and train our model in the Marian toolkit \citep{junczys-dowmunt-etal-2018-marian}. For the CTC implementation, we use the warp-ctc library\footnote{\url{https://github.com/baidu-research/warp-ctc}}. We release our code at \url{https://github.com/jindrahelcl/marian-dev}. The trained model (and a number of different variants including models in opposite translation direction) can be downloaded at \url{https://data.statmt.org/nar}.

\section{Results}
We refer the reader to the original paper for more details about the evaluation and its results. The model we submitted is denoted in the paper as ``large''. A summary of the results follows.

\paragraph{Translation Quality.} To summarize the main findings, the model achieves a competitive BLEU score \citep{papineni-etal-2002-bleu} on the WMT~14 news test set \citep{bojar-etal-2014-findings}, which serves as a comparison to other non-autoregressive models that use this test set as the de facto standard benchmark. When evaluated on the WMT~19 news test set, our model obtains BLEU of 47.8, and a COMET score \citep{rei-etal-2020-comet} of 0.1485. Compared to an similarly-sized autoregressive teacher model with 50.5 BLEU and COMET of 0.4110, we see a somewhat surprising gap between the COMET scores while BLEU scores are relatively close. We hypothesize that the errors that the non-autoregressive model makes are out of the training domain of the COMET models, which makes them more sensitive towards this kind of errors.

\paragraph{Decoding Time.}
We evaluated our models on the one million sentences benchmark used in the previous editions of this task \citep{heafield-etal-2021-findings}, and we tried to reproduce the official hardware setup to large extent. For CPU decoding, we measured time to translate the test set
on an Intel Xeon 6354 server from Oracle Cloud, with 36 cores. We run the evaluation only in the batch decoding mode, as the models were too slow to decode with a single sentence in batch. With the submitted model, the translation on CPU took 7,434 seconds (using batch of 16 sentences). 

We used a single Nvidia A100 GPU for GPU decoding. In the latency setup, the translation took 7,020 seconds, and the batched decoding ($b=128$) took 782 seconds. When compared with other submissions to this task, we find that the smallest difference is indeed found in the GPU decoding latency setting. However, the optimized models submitted to last year's round still achieved significantly better decoding times.

\section{Conclusions}
We submit a non-autoregressive system to the Efficient Translation Shared Task to the WMT~22. The model is trained with connectionist temporal classification, which allows the generation of empty tokens and thus making generation of sentences of various length possible while retaining the conditional independence among output tokens without explicit length estimation.

The main motivation of this submission is to provide a reasonable baseline system
for future research. We believe that the sub-field of non-autoregressive NMT cannot progress without controlled decoding speed evaluation, which is exactly what the shared task organizers provide.

\section*{Acknowledgements}
This project received funding from the European Union’s Horizon Europe Innovation programme under grant agreement no. 101070350 (HPLT).
Our work has been using data provided by the LINDAT/CLARIAH-CZ Research Infrastructure, supported by the Ministry of Education, Youth and Sports of the Czech Republic (Project No. LM2018101).

\bibliography{anthology,custom}
\bibliographystyle{acl_natbib}

\end{document}

%% file: img/non-autoregressive.tex
\def\inputsize{7}

\begin{tikzpicture}[]

\draw (\inputsize / 2 + 0.1, -0.1) node {Input token embeddings};

\foreach \i in {0,...,\inputsize} {
	\draw (\i,-0.5) rectangle (\i+0.2,-1);
    \draw [->] (\i+0.1,-1) -- (\i+0.1, -1.25);
};

\draw (0, -1.25) rectangle (\inputsize + 0.2, -2.25);
\draw (\inputsize / 2 + 0.1, -1.75) node {Encoder};

\foreach \i in {0,...,\inputsize} {
	\draw [->] (\i+0.1,-2.25) -- (\i+0.1, -2.5);
    \draw[fill=yellow!40] (\i,-2.5) rectangle (\i+0.2,-3);

    \draw [->] (\i+0.1,-3) -- (\i+0.1, -3.25);
	\draw[fill=blue!40] (\i,-3.25) rectangle (\i+0.2,-3.75);
	\draw[fill=red!40] (\i,-3.75) rectangle (\i+0.2,-4.25);

    \draw [dashed,->] (\i+0.1,-4.25) -  - (\i+0.1, -4.75);
    \draw [dashed,->] (\i+0.2,-3.5) .. controls (\i + 0.6, -3.65) .. (\i+0.6, -4.75);

	\draw[fill=red!40] (\i,-4.75) rectangle (\i+0.2,-5.25);
	\draw[fill=blue!40] (\i + 0.5,-4.75) rectangle (\i+0.7,-5.25);

    \draw [->] (\i+0.1,-5.25) - - (\i+0.1, -5.5);
    \draw [->] (\i+0.6,-5.25) - - (\i+0.6, -5.5);
};

\draw (\inputsize + 1.2, -2.75) node {$\mathbf{h}$};
\draw (\inputsize + 1.2, -3.75) node {$W_\text{spl}\mathbf{h}$};
\draw (\inputsize + 1.2, -5.00) node {$\mathbf{s}$};

\draw (0, -5.5) rectangle (\inputsize + 0.7, -6.5);
\draw (\inputsize / 2 + 0.5 + 0.1, -6.0) node {Decoder};

\draw [fill=green!80!black!60] (\inputsize / 2 + 0.4,-7.2) circle [x radius=\inputsize / 2 + 0.4, y radius=0.5];
\draw (\inputsize / 2 + 0.6, -7.2) node {Connectionist Temporal Classification};

\foreach \i in {0,...,\inputsize} {
   \draw [->] (\i+0.1,-7.9) - - (\i+0.1, -8.15);
   \draw [->] (\i+0.6,-7.9) - - (\i+0.6, -8.15);
}

\draw  (0+0.1,-8.4) node {$w_1$};
\draw  (0+0.6,-8.4) node {$w_2$};
\draw  (1+0.1,-8.4) node {$w_3$};
\draw  (1+0.6,-8.4) node {$\varnothing$};
\draw  (2+0.1,-8.4) node {$w_4$};
\draw  (2+0.6,-8.4) node {$\varnothing$};
\draw  (3+0.1,-8.4) node {$w_5$};
\draw  (3+0.6,-8.4) node {$w_6$};
\draw  (4+0.1,-8.4) node {$\varnothing$};
\draw  (4+0.6,-8.4) node {$\varnothing$};
\draw  (5+0.1,-8.4) node {$\varnothing$};
\draw  (5+0.6,-8.4) node {$w_7$};
\draw  (6+0.1,-8.4) node {$w_8$};
\draw  (6+0.6,-8.4) node {$\varnothing$};
\draw  (7+0.1,-8.4) node {$w_9$};
\draw  (7+0.6,-8.4) node {$\varnothing$};

\draw (\inputsize / 2 + 0.3, -8.95) node {Output tokens / null symbols};

\end{tikzpicture}